\documentclass[sigconf]{acmart}

\copyrightyear{2025}
\acmYear{2025}
\setcopyright{cc}
\setcctype{by}
\acmConference[CIKM '25]{Proceedings of the 34th ACM International Conference on Information and Knowledge Management}{November 10--14, 2025}{Seoul, Republic of Korea}
\acmBooktitle{Proceedings of the 34th ACM International Conference on Information and Knowledge Management (CIKM '25), November 10--14, 2025, Seoul, Republic of Korea}
\acmDOI{10.1145/3746252.3760796}
\acmISBN{979-8-4007-2040-6/2025/11}

\usepackage{balance}
\usepackage{multirow}
\usepackage{booktabs}
\usepackage{graphicx}
\usepackage{colortbl}
\usepackage{stfloats}
\usepackage{pifont}
\usepackage{array}
\usepackage[dvipsnames]{xcolor}
\usepackage[table,dvipsnames]{xcolor}
\usepackage[linesnumbered,ruled,vlined]{algorithm2e}
\usepackage[utf8]{inputenc}
\usepackage{tikz}
\usepackage{amsmath}
\usepackage{bm}
\definecolor{lightblue}{RGB}{220,230,241} 
\definecolor{lightteal}{RGB}{102, 194, 166}

\begin{document}

\title{FnRGNN: Distribution-aware Fairness in Graph Neural Network }

\author{SoYoung Park}
\orcid{0009-0000-7755-0041}
\affiliation{%
  \institution{Chungnam National University}
  \city{Daejeon}
  \country{Republic of Korea}
}
\email{sypark1452@o.cnu.ac.kr}

\author{Sungsu Lim}
\authornote{Corresponding author.}
\orcid{0000-0001-5924-3398}
\affiliation{%
  \institution{Chungnam National University}
  \city{Daejeon}
  \country{Republic of Korea}
}
\email{sungsu@cnu.ac.kr}

\renewcommand{\shortauthors}{SoYoung Park and Sungsu Lim}

\begin{abstract}
Graph Neural Networks (GNNs) excel at learning from structured data, yet fairness in regression tasks remains underexplored. Existing approaches mainly target classification and representation-level debiasing, which cannot fully address the continuous nature of node-level regression. We propose \texttt{FnRGNN}, a fairness-aware in-processing framework for GNN-based node regression that applies interventions at three levels: (i) structure-level edge reweighting, (ii) representation-level alignment via MMD, and (iii) prediction-level normalization through Sinkhorn-based distribution matching. This multi-level strategy ensures robust fairness under complex graph topologies. Experiments on four real-world datasets demonstrate that \texttt{FnRGNN} reduces group disparities without sacrificing performance. Code is available at \url{https://github.com/sybeam27/FnRGNN}.
\end{abstract}


  
\maketitle

\section{Introduction}
\label{sec:intro}
Graph Neural Networks (GNNs) are increasingly deployed in sensitive domains such as social networks~\cite{saxena2024fairsna}, recommendation systems~\cite{fu2020fairness}, and healthcare~\cite{wang2024faircare}. While fairness issues have been widely reported in general machine learning systems—such as gender-biased job recommendations~\cite{rus2022closing}, ad delivery favoring men~\cite{lambrecht2019algorithmic}, and racially skewed online exposure~\cite{sweeney2013discrimination}—similar risks exist in GNN-based models. GNNs, in particular, tend to amplify structural biases present in graphs through their message-passing mechanisms, making them especially prone to reinforcing disparities related to sensitive attributes like race and gender~\cite{lin2024bemap, jiang2024chasing, laclau2022survey, dong2022structural}. Since such attributes can be implicitly encoded during training, ensuring fairness requires interventions at the algorithmic level—beyond input pre-processing or post-hoc correction~\cite{ling2023learning, ma2022learning}.

Despite growing interest in in-processing methods for fair GNN training, most existing approaches address bias at only one stage, most often the representation level, by removing sensitive information from node embeddings~\cite{dai2021say, zhu2023fair, agarwal2021towards}. However, this overlooks structural bias propagated through message passing. Structure-level debiasing methods have been proposed~\cite{jiang2024chasing, loveland2022fairedit}, but they are largely tailored for classification and struggle to generalize to regression.
Node-level regression plays a critical role in applications such as disease risk estimation~\cite{li2020graph}, recidivism prediction~\cite{zhou2024hdm}, and credit risk modeling~\cite{balmaseda2023predicting}, yet fairness in graph-based regression remains underexplored. Unlike classification or link prediction, where fairness is typically defined over discrete labels, regression involves continuous targets, making accuracy alone insufficient. Ensuring equitable outcomes requires evaluating and enforcing distributional parity across sensitive groups~\cite{zhang2024regexplainer, chen2024fairness}. Even small prediction disparities can lead to significant unfairness in high-stakes domains, highlighting the need for a unified, distribution-aware framework that mitigates bias across structural, representational, and predictive stages.

To address this gap, we propose \textbf{\texttt{FnRGNN}}, a fairness-aware in-processing framework for node-level regression. Standard GNN regressors often produce disparate output distributions across sensitive groups. \texttt{FnRGNN} mitigates these disparities through a distribution-aware design that enforces fairness throughout training. It integrates three components:
(i) \textit{Structure-level} edge reweighting to suppress bias propagation during message passing~\cite{zhu2023fair, lin2024bemap},
(ii) \textit{Representation-level} alignment to reduce differences in the embedding space~\cite{gretton2006kernel}, and
(iii) \textit{Prediction-level} normalization to match output distributions across groups via full-shape alignment~\cite{feydy2019interpolating, li2015generative}.
This unified, multi-level framework extends fair GNN research beyond classification, tackling fairness in continuous prediction where accuracy alone is insufficient. Our contributions are as follows:
\begin{enumerate}
\item \textbf{Fairness in Graph Node Regression:} We focus on node-level regression, an important yet underexplored setting in fair GNN, and address its unique fairness challenges from predicting continuous outcomes across diverse groups.
\item \textbf{Multi-level Distribution-aware Framework:} We introduce \texttt{FnRGNN}, a distribution-aware framework operating across structure, representation, and prediction levels.
\item \textbf{Empirical Validation:} Experiments on four real-world datasets demonstrate that \texttt{FnRGNN} effectively reduces group disparities while preserving high predictive accuracy.
\end{enumerate}

\section{Related Work}
\label{sec:rel_work}
GNNs tend to amplify biases in graph-structured data~\cite{laclau2022survey, dong2022structural}, making fairness a critical concern in research. Recent studies focus on in-processing methods that mitigate bias during training, categorized into structure, representation, and prediction levels~\cite{chen2024fairness}.

\textbf{(i) \textit{Structure-level}} methods mitigate bias by modifying graph topology or aggregation mechanisms. These include adjusting message passing schemes~\cite{jiang2024chasing}, editing the graph structure~\cite{loveland2022fairedit, dong2022edits, dong2023interpreting}, reweighting or resampling~\cite{lin2024bemap, liu2023generalized}, and applying debiasing techniques~\cite{buyl2020debayes}. Other approaches leverage counterfactual reasoning~\cite{ma2022learning} or multi-level models~\cite{he2023fairmile} to account for structural sources of bias.

\textbf{(ii) \textit{Representation-level}} methods focus on learning node embeddings that are invariant to sensitive attributes. Common strategies include adversarial training~\cite{dai2021say}, distribution alignment~\cite{zhu2023fair}, orthogonalization~\cite{palowitch2020debiasing}, and normalization techniques~\cite{agarwal2021towards}. 

\textbf{(iii) \textit{Prediction-level}} methods directly target output fairness using ranking-based objectives~\cite{dong2021individual}, fairness-regularized losses~\cite{kang2022rawlsgcn}, or output distribution alignment via regularization~\cite{gretton2006kernel}.

While prior work has advanced fairness in GNNs, most methods target classification or link prediction and offer limited support for node-level regression, where prediction targets are continuous and biases arise more subtly. Fairness in regression tasks requires more than embedding-level invariance; it must account for structural bias and align output distributions across groups~\cite{zhang2024regexplainer}, calling for a holistic approach across structure, representation, and prediction.

\section{Preliminaries}
\label{sec:preli}
\paragraph{\textbf{Graph Neural Networks}} 
We consider a graph \( G = (\mathcal{V}, \mathcal{E}) \), where \( \mathcal{V} \) is the set of nodes and \( \mathcal{E} \subseteq \mathcal{V} \times \mathcal{V} \) is the set of edges. Each node \( i \in \mathcal{V} \) is associated with a feature vector \( \mathbf{x}_i \in \mathbb{R}^d \), and the complete feature matrix is denoted by \( \mathbf{X} \in \mathbb{R}^{n \times d} \). Each node also has a continuous regression target \( y_i \in \mathbb{R} \), forming the label vector \( \mathbf{y} \in \mathbb{R}^n \).
A GNN learns node representations by aggregating neighborhood information through stacked message-passing layers. At each layer \( l \), the representation of node \( i \) is updated as
\begin{equation}
    \mathbf{h}_i^{(l)} = \sigma \left( W^{(l)} \cdot \text{AGG} \left( \{ \mathbf{h}_u^{(l-1)} : u \in \mathcal{N}(i) \} \right) \right),
\end{equation}
where \( \mathcal{N}(i) \) denotes the set of neighbors of node \( i \), \( \text{AGG} \) is an aggregation function (e.g., mean, sum, or attention), and \( \sigma \) is a non-linear activation. The final node embedding \( \mathbf{h}_i^{(L)} \) is then passed to a regression head to produce the prediction \( \hat{y}_i \in \mathbb{R} \).

\paragraph{\textbf{Fairness in Node Regression}}  
We define fairness as consistency of predictions across groups defined by a binary sensitive attribute \(s_i \in \{0,1\}\).  
We adopt two criteria: (1) \textit{mean parity} — small difference in group-wise expected predictions,  
\( \left| \mathbb{E}[\hat{y}_i \mid s_i = 0] - \mathbb{E}[\hat{y}_i \mid s_i = 1] \right| \approx 0\);  
(2) \textit{distributional parity} — small divergence between group-wise predictive distributions,  
\(\mathbb{D}( P_{\hat{y}_i \mid s_i = 0}, P_{\hat{y}_i \mid s_i = 1} ) \approx 0\).  

\section{Methodology}
\label{sec:methodology}
\begin{figure*}[t]
    \centering
    \includegraphics[width=0.90\linewidth]{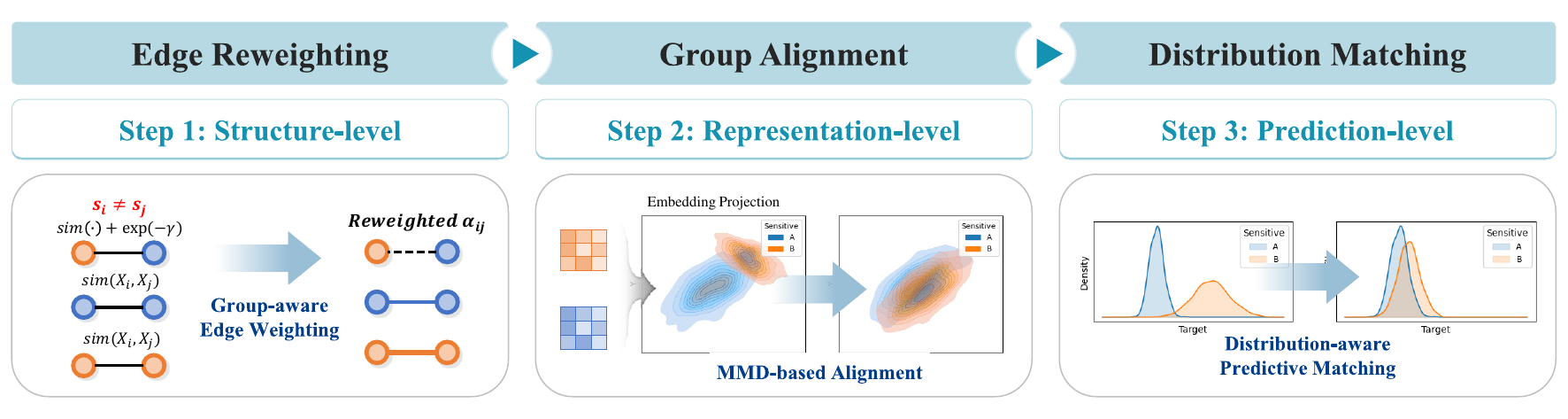}
    \Description{Overview of \texttt{FnRGNN}, which integrates fairness interventions at three levels: (i) structure, (ii) representation, and (iii) prediction. Each component addresses group-level disparities in node-level regression, as introduced in Sec.~\ref{sec:methodology}.}
    \vspace{-0.2cm}
    \caption{Overview of \texttt{FnRGNN}, which integrates fairness interventions at three levels: (i) structure, (ii) representation, and (iii) prediction. Each component addresses group-level disparities in node-level regression, as introduced in Sec.~\ref{sec:methodology}.
    }
    \label{fig:method}
\end{figure*}
We propose \texttt{FnRGNN}, a fairness-aware graph neural network for node-level regression. Its goal is to minimize prediction error while promoting fairness across sensitive groups, as defined in Sec.~\ref{sec:preli}. To achieve this, \texttt{FnRGNN} incorporates fairness interventions at three levels: (i) \textit{structure} (Sec.~\ref{subsec:step1}), (ii) \textit{representation} (Sec.~\ref{subsec:step2}), and (iii) \textit{prediction} (Sec.~\ref{subsec:step3}), as illustrated in Fig.~\ref{fig:method}.

\subsection{Edge Reweighting for Fair Structure}
\label{subsec:step1}
To prevent the amplification of bias through message passing~\cite{hamilton2017inductive, kipf2016semi}, we propose a hybrid edge weighting that jointly considers node feature similarity and group demographic dissimilarity. 
For each edge \( (i, j) \in \mathcal{E} \), we define the adjusted edge weight \( \alpha_{ij} \) as:
\begin{equation}
    \alpha_{ij} = \text{sim}(\mathbf{x}_i, \mathbf{x}_j) \cdot \exp\left(-\gamma \cdot \mathbb{I}[s_i \neq s_j] \right),
\end{equation}
where \( \text{sim}(\cdot, \cdot) \) captures feature-level closeness between node pairs, while the indicator function \( \mathbb{I}[\cdot] \) checks whether their sensitive attributes differ. An exponential penalty controlled by hyperparameter \( \gamma \) reduces the weights of cross-group edges, mitigating bias without hard pruning. This soft reweighting preserves graph connectivity and is directly integrated into standard GNN propagation for fair representation learning~\cite{zhu2023fair, lin2024bemap}.
At each layer \( l \), node embeddings are updated as:
\begin{equation}
    \mathbf{h}_i^{(l+1)} = \sigma \left( \sum_{j \in \mathcal{N}(i)} \alpha_{ij} \cdot \mathbf{h}_j^{(l)} \mathbf{W}^{(l)} \right),    
\end{equation} 
where \( \mathbf{W}^{(l)} \) is a learnable transformation matrix and \( \sigma(\cdot) \) is a non-linear activation function. By integrating fairness-aware edge weights directly into the message passing process, our model effectively modulates structural bias during representation learning.

\subsection{Group Alignment for Fair Representation}
\label{subsec:step2}
To mitigate representational bias, we introduce a kernel-based regularization using Maximum Mean Discrepancy (MMD)~\cite{gretton2006kernel}. Given a binary sensitive attribute \( s_i \), we define group \( \mathcal{G}_a = \{ i \in \mathcal{V} \mid s_i = a \} \) for \( a \in \{0, 1\} \). MMD encourages the embedding distributions of these groups to be statistically aligned—not only in terms of the mean but also higher-order moments—thus promoting fairer representations. The MMD loss is computed as:
\begin{align}
\mathcal{L}_{\text{MMD}} &= 
\mathbb{E}_{\mathbf{x}, \mathbf{x}' \in \mathcal{G}_0}[k(\mathbf{x}, \mathbf{x}')] 
+ \mathbb{E}_{\mathbf{y}, \mathbf{y}' \in \mathcal{G}_1}[k(\mathbf{y}, \mathbf{y}')] \nonumber \\
&\quad - 2 \mathbb{E}_{\mathbf{x} \in \mathcal{G}_0, \mathbf{y} \in \mathcal{G}_1}[k(\mathbf{x}, \mathbf{y})],
\end{align}
where \( k(\cdot, \cdot) \) is a positive-definite kernel function, typically instantiated as the Gaussian RBF kernel~\cite{scholkopf1997comparing}, defined by \( k(\mathbf{X}, \mathbf{y}) = \exp\left(-\|\mathbf{X} - \mathbf{y}\|^2 / (2\sigma^2)\right) \), to measure similarity.
Minimizing \( \mathcal{L}_{\text{MMD}} \) encourages the model to learn group-invariant representations by reducing distributional discrepancies in embedding space. This regularization serves as a soft fairness constraint and integrates seamlessly with the main regression objective.

\subsection{Distribution Matching for Fair Prediction}
\label{subsec:step3}
To ensure output-level fairness, we apply a dual regularization strategy that aligns global distributions and preserves local statistical consistency across sensitive groups. Unlike standard moment matching, our approach accounts for higher-order differences in the output space.
To address this, we combine moment-based matching with Sinkhorn divergence~\cite{feydy2019interpolating}, a smoothed version of optimal transport (OT)~\cite{vincent2022template} that measures the geometric cost of transforming one distribution into another under entropic regularization.
Specifically, the output-level fairness loss is defined as \(\mathcal{L}_{\text{dist}} = \mathcal{L}_{\text{sinkhorn}} + \mathcal{L}_{\text{moment}},\) where the Sinkhorn term is given by:
% \begin{equation}
% \mathcal{L}_{\text{sinkhorn}} = OT_\varepsilon(\hat{\mathbf{y}}_{\mathcal{G}_0}, \hat{\mathbf{y}}_{\mathcal{G}_1}) 
% - \frac{1}{2} \left[ OT_\varepsilon(\hat{\mathbf{y}}_{\mathcal{G}_0}, \hat{\mathbf{y}}_{\mathcal{G}_0}) 
% + OT_\varepsilon(\hat{\mathbf{y}}_{\mathcal{G}_1}, \hat{\mathbf{y}}_{\mathcal{G}_1}) \right],
% \end{equation}
\begin{align}
\mathcal{L}_{\text{sinkhorn}} &= 
OT_\varepsilon(\hat{\mathbf{y}}_{\mathcal{G}_0}, \hat{\mathbf{y}}_{\mathcal{G}_1}) \notag \\
&\quad - \tfrac{1}{2} \left[ OT_\varepsilon(\hat{\mathbf{y}}_{\mathcal{G}_0}, \hat{\mathbf{y}}_{\mathcal{G}_0}) 
+ OT_\varepsilon(\hat{\mathbf{y}}_{\mathcal{G}_1}, \hat{\mathbf{y}}_{\mathcal{G}_1}) \right],
\end{align}
with \( OT_\varepsilon \) denoting the entropic OT distance computed with regularization parameter \( \varepsilon \). This formulation stabilizes training by penalizing over-concentration and sampling noise.
In parallel, we apply a moment-based regularizer~\cite{li2015generative} to enforce consistency in the first and second moments:
\begin{equation}
\mathcal{L}_{\text{moment}} = 
\left| \mathbb{E}[\hat{\mathbf{y}}_{\mathcal{G}_0}] - \mathbb{E}[\hat{\mathbf{y}}_{\mathcal{G}_1}] \right| + \left| \text{Var}[\hat{\mathbf{y}}_{\mathcal{G}_0}] - \text{Var}[\hat{\mathbf{y}}_{\mathcal{G}_1}] \right|.
\end{equation}
This hybrid approach ensures robust fairness by aligning predictions and preserving group-level statistics.

\subsection{Training Procedure}
\begin{algorithm}[t]
\caption{Training of \texttt{FnRGNN}}
\label{alg:fnrgnn}
\KwIn{Graph $\mathcal{G}$, features $\mathbf{X}$, target $\mathbf{y}$, sensitive attribute $s$}
\KwOut{Trained model parameters $\theta$}
Initialize model parameters $\theta$ \;
\For{each epoch}{
  \textbf{1. Structure-level:} \\
  $\alpha_{ij} \!\leftarrow\! \mathrm{sim}(\mathbf{x}_i,\mathbf{x}_j) e^{-\gamma \mathbb{I}[s_i \neq s_j]}$ ; $h^{(2)} \leftarrow \text{GCNLayer}(X, A_\alpha)$ \\
  \textbf{2. Representation-level:} 
  $\mathcal{L}_{\mathrm{MMD}} \!=\! \mathrm{MMD}(\mathbf{H}^{(2)}_{\mathcal{G}_0}, \mathbf{H}^{(2)}_{\mathcal{G}_1})$ \\
  \textbf{3. Prediction-level:} 
  $\mathcal{L}_{\mathrm{dist}} \!=\! \mathcal{L}_{\mathrm{sinkhorn}} \!+\! \mathcal{L}_{\mathrm{moment}}$ \\
  \textbf{4. Optimize:} 
  $\mathcal{L}_{\mathrm{total}} \!=\! \mathrm{MSE} + \lambda_{\mathrm{MMD}}\mathcal{L}_{\mathrm{MMD}} + \lambda_{\mathrm{dist}}\mathcal{L}_{\mathrm{dist}}$
}
\end{algorithm}

In Alg.~\ref{alg:fnrgnn}, \texttt{FnRGNN} is trained in four stages. \textbf{Step 1} constructs a fairness-aware graph by reweighting edges with a similarity-based coefficient $\alpha_{ij}$ that penalizes cross-group connections. The resulting adjacency matrix is used in a GCN to generate hidden representations, followed by an MLP that outputs predictions $\hat{\mathbf{y}}$. \textbf{Step 2} promotes representation-level fairness by minimizing the MMD loss $\mathcal{L}_{\text{MMD}}$, aligning embeddings across sensitive groups. \textbf{Step 3} ensures prediction-level fairness via distributional alignment using $\mathcal{L}_{\text{dist}}$, which combines Sinkhorn distance and moment matching. \textbf{Step 4} defines the total loss $\mathcal{L}_{\text{total}}$ as the sum of prediction error and fairness terms. Model parameters $\theta$ are optimized via Adam, with fairness losses computed per mini-batch for stability.

\section{Experiments}
\label{sec:experiment}
\begin{table}[t]
\centering
\caption{
Dataset statistics and label distributions by sensitive group. 
The rightmost column shows target value differences across sensitive attributes.
}
\vspace{-0.2cm}
\resizebox{0.47\textwidth}{!}{
\begin{tabular}{
    cc|
    >{\centering\arraybackslash}p{1cm} 
    >{\centering\arraybackslash}p{1cm} 
    >{\centering\arraybackslash}p{1cm} 
    |
    >{\centering\arraybackslash}p{1.3cm}
}
\toprule
\textbf{Dataset} & \textbf{Group $\mathcal{G}$} & \textbf{$|\mathcal{V}|$} & \textbf{$|\mathcal{E}|$} & \textbf{$\dim(\mathbf{X})$} & \textbf{Dist.} \\
\midrule
Pokec-z & Region & 67,796 & 1,303,712 & 276 & \raisebox{-0.5\height}{\includegraphics[width=\linewidth]{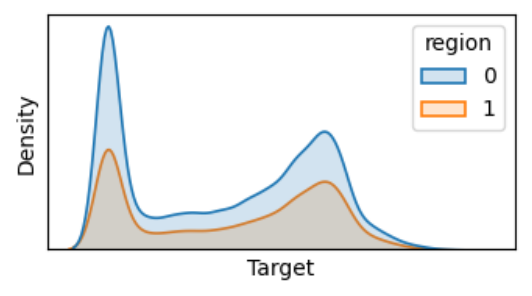}}\\
\midrule
Pokec-n & Gender & 66,569 & 1,100,663 & 265 & \raisebox{-0.5\height}{\includegraphics[width=\linewidth]{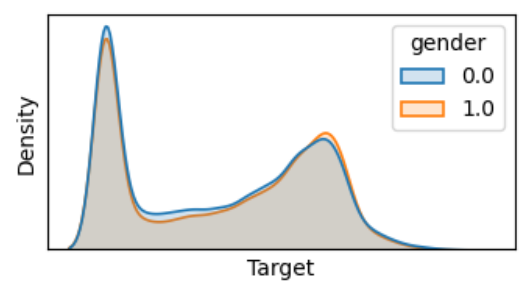}}\\
\midrule
NBA & Country & 403   & 19,357   & 95  & \raisebox{-0.5\height}{\includegraphics[width=\linewidth]{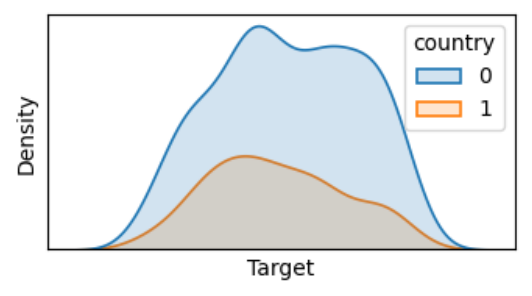}} \\
\midrule
German & Gender & 1,000  & 44,484   & 29  & \raisebox{-0.5\height}{\includegraphics[width=\linewidth]{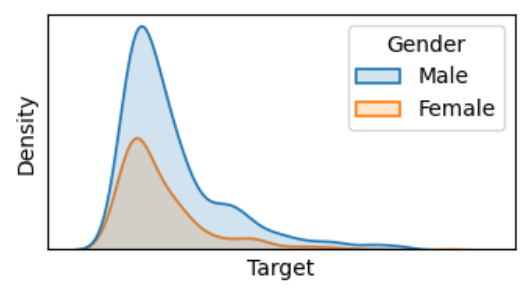}} \\
\bottomrule
\end{tabular}
}
\label{tab:dataset}
\end{table}

\subsection{Experiment Settings}
\paragraph{\textbf{Datasets}}  
We evaluate on four real-world graph datasets with diverse structures and sensitive attributes (Tab.\ref{tab:dataset}):
\textbf{Pokec-z} and \textbf{Pokec-n}\cite{takac2012data} are Slovak social network subsets for profile completion; Pokec-z shows mean-shift bias, while Pokec-n has near-identical gender distribution.
\textbf{NBA}\cite{dai2021say} contains basketball data for predicting minutes per game, with nationality as the sensitive attribute; despite group size imbalance, distributions are similar.
\textbf{German}\cite{agarwal2021towards} is credit data with gender as the sensitive attribute; group distributions differ in shape and mean, indicating moderate bias.

\paragraph{\textbf{Baselines and Implementation}}
We compare \texttt{FnRGNN} with representative fair GNN baselines, categorized into three fairness strategies levels: (i) structure — FMP~\cite{jiang2024chasing}, EDITS~\cite{dong2022edits}; (ii) representation — FairGNN~\cite{dai2021say}, GMMD~\cite{zhu2023fair}; and (iii) prediction — REDRESS~\cite{dong2021individual}, and GCN$_{mean}$, a GCN-based regressor that performs group-wise mean matching.
All models were implemented in PyTorch Geometric, adapting regression by replacing classification loss with MSE. Training used Adam optimizer for 500 epochs with early stopping, and results were averaged over five runs. Unless noted, we used a 2-layer GCN~\cite{kipf2016semi} encoder (64 hidden units, learning rate \(10^{-3}\), weight decay \(10^{-5}\)). \texttt{FnRGNN} extends this base with (i) MMD~\cite{gretton2006kernel} alignment, (ii) Sinkhorn regularization~\cite{feydy2019interpolating}, and (iii) edge reweighting. Hyperparameters \((\lambda_{\text{MMD}}, \lambda_{\text{dist}}, \gamma)\) were tuned via NSGA-II in Optuna, with losses computed from 500 sampled nodes per group.

\paragraph{\textbf{Evaluation Metrics}}  
To evaluate fairness in regression, we adopt group-wise metrics suitable for continuous outputs, since classification-based notions (e.g., demographic parity) are not directly applicable~\cite{chen2024fairness}. Given a binary sensitive attribute \( s_v \in \{0,1\} \), we define groups \( \mathcal{G}_0 \) and \( \mathcal{G}_1 \) and compute:  
(i) Mean Gap (\textbf{MG}) \(= \left| \bar{\mathbf{y}}_{\mathcal{G}_0} - \bar{\mathbf{y}}_{\mathcal{G}_1} \right| \), (ii) Variance Gap (\textbf{VG}) \(= \left| \text{Var}(\hat{\mathbf{y}}_{\mathcal{G}_0}) - \text{Var}(\hat{\mathbf{y}}_{\mathcal{G}_1}) \right| \), and  (iii) Wasserstein Distance (\textbf{WD})~\cite{villani2021topics}, which quantifies distributional divergence as 
\( \inf_{\gamma \in \Pi} \mathbb{E}_{(i,j) \sim \gamma} [ |\hat{y}_i - \hat{y}_j| ] \), where \( \gamma \) denotes a transport plan between group-wise prediction distributions.
Together, MG and VG assess moment disparities, while WD captures full-shape differences in prediction distributions. 
For predictive accuracy, we report Mean Squared Error (\textbf{MSE}) \( = \frac{1}{n} \sum_{i=1}^{n}(\hat{y}_i - y_i)^2 \) and Mean Absolute Error (\textbf{MAE}) \( = \frac{1}{n} \sum_{i=1}^{n} |\hat{y}_i - y_i| \).

\begin{table*}[t]
\centering
\renewcommand{\arraystretch}{1.15}
\caption{
Results on four datasets (average of five runs). 
MSE/MAE assess accuracy; MG, VG, and WD assess fairness. 
(i) Structure-level: FMP~\cite{jiang2024chasing}, EDITS~\cite{dong2022edits}; (ii) Representation-level: FairGNN~\cite{dai2021say}, GMMD~\cite{zhu2023fair}; (iii) Prediction-level: REDRESS~\cite{dong2021individual}, GCN$_{mean}$. 
EDITS and REDRESS are omitted for Pokec datasets due to GPU memory limits.
}
\vspace{-0.2cm}
\resizebox{1.0\textwidth}{!}{
\begin{tabular}{c|l|ccccc||c|ccccccc}
\toprule
\textbf{Metrics (\textcolor{blue}{\bm{$\downarrow$}})} & \textbf{Dataset}
& \textbf{FMP} & \textbf{FairGNN} & \textbf{GMMD} & \textbf{GCN} & \textbf{FnRGNN}
& \textbf{Dataset} & \textbf{FMP} & \textbf{EDITS} & \textbf{FairGNN} & \textbf{GMMD} & \textbf{REDRESS} & \textbf{GCN} & \textbf{FnRGNN}
\\

\midrule
MSE & \multirow{5}{*}{Pokec-z} 
& 0.3178 & 0.6921 & \textbf{0.0342} &  0.4333 & \underline{0.0622} 
& \multirow{5}{*}{NBA} & 1.0001 & 746.28 & 1.1417 & \underline{0.3496} & 0.9977 & 1.6414 & \textbf{0.1495}
\\
MAE & & 0.4986 &  0.7052 & \textbf{0.1247} &  0.5040 & \underline{0.1829} 
& & 0.8570 & 21.97 & 0.8177 & \underline{0.3803} & 0.8555 & 0.8016 & \textbf{0.3024}
\\
MG &  & 0.0269 &  0.0416 & 0.0747 &  \textbf{0.0084} & \underline{0.0171} 
& & \underline{0.0019} & 13.80 & 0.2204 & 0.1156 & \textbf{0.0001} & 0.5501 & 0.0415
\\
VG &  & \textbf{0.0026} &  0.0392 & 0.0150 &  0.1074 & \underline{0.0046} 
& & \underline{0.0003} & 216.40 & 0.3509 & 0.2891 & \textbf{0.0000} & 4.3050 & 0.1876
\\
WD &  & 0.0063 &  \underline{0.0055} & \textbf{0.0028} &  0.0604 & \underline{0.0055} 
& & \textbf{0.0038} & 13.64 & 0.0185 & 0.1085 & 0.0106 & 0.5279 & \underline{0.0070}
\\

\midrule
MSE & \multirow{5}{*}{Pokec-n} & 0.4857 & 0.5877 & \textbf{0.0301} & 0.4439 & \underline{0.0511} 
& \multirow{5}{*}{German} & 1.0934 & 304.53 & \underline{1.0141} & 1.7481 & 1.1700 & 1.0490 & \textbf{0.6994}
\\
MAE & & 0.6217 & 0.6427 & \textbf{0.1140} & 0.5040 & \underline{0.1557} 
& & 0.9511 & 13.05 & 0.8803 & 1.0308 & 0.8993 & \underline{0.8550} & \textbf{0.6995}
\\
MG & & 0.0232 & \underline{0.0064} & 0.0437 & \textbf{0.0008} & 0.0133 
& & \underline{0.0007} & 3.2408 & 0.1643 & 0.1812 & \textbf{0.0001} & 0.0010 & 0.1421
\\
VG & & \underline{0.0012} & 0.0156 & 0.0242 & 0.0046 & \textbf{0.0003} 
& & \underline{0.0001} & 28.98 & 0.0012 & 0.3665 & \textbf{0.0000} & 0.0250 & 0.0730
\\
WD & & 0.0143 & 0.0312 & \underline{0.0107} & 0.0240 & \textbf{0.0018} 
& & \underline{0.0465} & 0.7483 & 0.1090 & 0.0881 & 0.0662 & 0.0574 & \textbf{0.0217}
\\
\bottomrule
\end{tabular}
}
\label{tab:results_exp}
\end{table*}

\begin{table}[t]
\centering
\small
\caption{
Ablation results on four datasets: 
Case 1 — without edge reweighting; 
Case 2 — without group embedding alignment; 
Case 3 — with mean-only distribution matching.
}
\vspace{-0.2cm}
\resizebox{0.48\textwidth}{!}{
\begin{tabular}{c|c|cccc|c}
\toprule
\multirow{1}{*}{\textbf{Metric (\textcolor{blue}{\bm{$\downarrow$}})}} & \multirow{1}{*}{\textbf{Dataset}} & \multirow{1}{*}{\textbf{Vanilla}} & \textbf{Case 1} & \textbf{Case 2} & \textbf{Case 3} & \multirow{1}{*}{\textbf{Full}} \\
\midrule
MSE & \multirow{2}{*}{Pokec-z} & 0.4676 & 0.4646 & \underline{0.0846} & 0.4333 & \textbf{0.0622} \\
WD & & 0.0206 & \textbf{0.0053} & 0.0112 & 0.0604 & \underline{0.0055} \\

\midrule
MSE & \multirow{2}{*}{Pokec-n}& 0.4837 & 0.4417 & \underline{0.0789} & 0.4439 & \textbf{0.0511} \\
WD & & 0.0181 & 0.0123 & \underline{0.0036} & 0.0240 & \textbf{0.0018} \\

\midrule
MSE & \multirow{2}{*}{NBA} & 5.3053 & 0.5787 & 0.1695 & \underline{0.1641} & \textbf{0.1495} \\
WD & & 1.0585 & \underline{0.0321} & 0.0561 & 0.5279 & \textbf{0.0070} \\

\midrule
MSE & \multirow{2}{*}{German} & 1.4856 & 0.8654 & \underline{0.8072} & 1.0490 & \textbf{0.6994} \\
WD & & 0.0372  & \underline{0.0284} & 0.0524 & 0.0574 & \textbf{0.0217} \\

\bottomrule
\end{tabular}

}
\label{tab:ablation}
\end{table}

\subsection{Experiment Results}
We compare \texttt{FnRGNN} with representative fair GNN baselines on four datasets with varying graph structures and demographic imbalances.
As shown in Tab.~\ref{tab:results_exp}, \texttt{FnRGNN} consistently achieves a favorable trade-off between prediction accuracy and fairness across all three evaluation levels.
At the (i) \textit{structure-level}, \texttt{FnRGNN} clearly outperforms FMP and EDITS, especially on real-world datasets such as NBA and German. While FMP and EDITS suffer from high MSE and fairness gaps, \texttt{FnRGNN} effectively leverages edge reweighting to enhance both accuracy and group parity.
At the (ii) \textit{representation-level}, GMMD and FairGNN partially reduce fairness gaps, but often at the cost of prediction accuracy. For example, GMMD achieves low MSE on Pokec but fails to minimize group disparities (e.g., MG, WD). In contrast, \texttt{FnRGNN} achieves both low error and consistently small group-wise gaps, suggesting stronger alignment in the latent space.
At the (iii) \textbf{prediction-level}, \texttt{FnRGNN} significantly outperforms REDRESS and GCN$_{\text{mean}}$, especially on datasets like German and NBA. While REDRESS struggles with fairness, GCN$_{\text{mean}}$ underperforms in accuracy. \texttt{FnRGNN} achieves the lowest MSE and MAE while maintaining strong fairness scores across MG, VG, and WD.
These results demonstrate that \texttt{FnRGNN}'s multi-level design enables robust performance under distributional imbalance, jointly addressing structural, representational, and predictive biases.

\subsection{Ablation Study}
We conduct ablation studies on four regression datasets to evaluate the contribution of each component in \texttt{FnRGNN}. As shown in Tab.~\ref{tab:ablation}, the full model consistently achieves the best trade-off between MSE and WD.
Case 1 occasionally lowers WD but results in higher MSE, suggesting that this module enhances training stability and overall predictive accuracy.
Case 2 leads to a notable increase in MSE, especially on the Pokec datasets, emphasizing the importance of alignment in the embedding space for both fairness and accuracy. Case 3 increases WD in most datasets, underscoring the necessity of aligning full output distributions to mitigate bias effectively. Overall, these results confirm that the three components of \texttt{FnRGNN} are complementary and jointly contribute to fairness in regression.

\section{Conclusion}
\label{sec:con}
We present \texttt{FnRGNN}, a fairness-aware GNN for node regression that applies multi-level interventions to mitigate bias. Our framework integrates edge reweighting for fair message passing, MMD-based representation alignment, and Sinkhorn-augmented distribution matching for prediction fairness. Experiments on real-world datasets show that each module improves fairness without compromising predictive accuracy. These results demonstrate the effectiveness of structured fairness regularization without the need for adversarial training or complex architectures. Future work will extend \texttt{FnRGNN} to multi-class and dynamic graph scenarios and investigate theoretical fairness guarantees under distribution shifts.

\begin{acks}
This work was supported by the National Research Foundation of Korea (NRF) grant funded by the Korea government (MSIT) (No. RS-2023-00214065) and by Institute of Information \& Communications Technology Planning \& Evaluation (IITP) grant funded by the Korea government (MSIT) (No. RS-2022-00155857, Artificial Intelligence Convergence Innovation Human Resources Development (Chungnam National University)).
\end{acks}

\newpage
\section*{Generative AI Usage Disclosure}
This paper made limited and transparent use of generative AI tools—specifically ChatGPT, Grok, and Grammarly—for the following purposes: (i) minor language editing, including grammar, spelling, and clarity improvements; (ii) preliminary exploration of research directions; (iii) verification and clarification of technical concepts to support the author's understanding; and (iv) assistance with LaTeX formatting. No AI-generated text was used verbatim, and all core ideas, arguments, and interpretations are solely the author’s own. AI tools were used strictly as assistive technologies and did not influence the conceptual development or analytical reasoning of the work.

\bibliographystyle{ACM-Reference-Format}
\balance
\bibliography{reference}

\end{document}